\documentclass[sigconf, 10pt]{acmart}

\AtBeginDocument{%
  }

\setcopyright{acmlicensed}

\acmYear{2025}\copyrightyear{2025}
\acmConference[SEC '25]{The Tenth ACM/IEEE Symposium on Edge Computing}{December 3--6, 2025}{Arlington, VA, USA}
\acmBooktitle{The Tenth ACM/IEEE Symposium on Edge Computing (SEC '25), December 3--6, 2025, Arlington, VA, USA}
\acmDOI{10.1145/3769102.3770611}
\acmISBN{979-8-4007-2238-7/25/12}

\usepackage{algpseudocode}
\usepackage{graphicx}
\usepackage{textcomp}
\usepackage{xcolor}
\usepackage{algorithm}
\usepackage{algorithmicx}
\usepackage{multicol}
\usepackage{multirow}

\newtheorem{definition}{\textbf{Definition}}  

\newcommand\ourmethod{\textsc{FrameScope}}

\begin{document}

\title[\ourmethod{}]{\ourmethod{}: Temporal Data Valuation for Stream Active Learning in Autonomous Vehicle Systems}

\author{Yuheng Zhu}
\email{yzhu63@ncsu.edu}
\affiliation{%
  \institution{Department of Computer Science\\North Carolina State University}
  \city{Raleigh}
  \state{NC}
  \country{USA}}

\author{Man-Ki Yoon}
\email{man-ki.yoon@ncsu.edu}
\affiliation{%
  \institution{Department of Computer Science\\North Carolina State University}
  \city{Raleigh}
  \state{NC}
  \country{USA}}

\renewcommand{\shortauthors}{Zhu and Yoon}

\begin{abstract}
Autonomous vehicles operate in dynamic, ever-changing environments where new scenarios and edge cases constantly emerge. As a result, static learning models are inadequate for ensuring safe and reliable operation. Continuous learning is essential for adapting to these evolving conditions and maintaining robust performance across diverse real-world settings. However, autonomous vehicles generate massive streams of visual data during operation, and existing continuous learning approaches typically rely on heuristic sampling methods that fail to capture temporal dynamics, often overlooking critical learning opportunities or selecting redundant frames. In this paper, we introduce \ourmethod{}, a temporal data valuation framework for continuous learning in autonomous vehicles. \ourmethod{} extends neural tangent kernel theory to temporal domains, enabling principled valuation of streaming visual data. Unlike cloud-centric methods that transmit all video data for processing, our approach performs principled, local frame selection on the vehicle and queries a cloud-based oracle model only for labels of those high-value frames. Extensive experiments across multiple domain shifts show that \ourmethod{} consistently outperforms existing methods, achieving higher sample efficiency and significantly reducing catastrophic forgetting in autonomous vehicle perception. By valuing data on the vehicle and querying only labels for selected frames, \ourmethod{} reduces bandwidth requirements, enabling scalable operation with a lightweight cloud labeling service.
\end{abstract}



\maketitle
\section{Introduction}

Autonomous vehicles require reliable computer vision systems that can accurately perceive and interpret complex driving environments. A key challenge in deploying these models is ensuring robust generalization to novel driving scenarios and diverse environmental conditions. The fundamental difficulty stems from the inherent dynamism of real-world environments; autonomous vehicles frequently encounter unexpected situations, such as unfamiliar geographic regions, that may be underrepresented in their initial training data. This creates the critical need for \emph{continuous learning}~\cite{thrun1995lifelong} capabilities that enable perception models to adapt over time while avoiding catastrophic forgetting of previously learned knowledge.

A key bottleneck in continuous learning is the principled \emph{data selection} from the massive volumes of visual data streaming generated by modern autonomous vehicles. Although these vehicles can produce thousands of frames per minute, not all frames contribute equally to model improvement; many contain redundant information, whereas others capture critical edge cases invaluable for learning. The challenge is to develop methods that effectively identify valuable frames while maintaining computational efficiency.

Existing data selection approaches often rely on heuristic methods, such as random sampling, frame-based temporal sampling, or uncertainty-based selection~\cite{shorten2019survey, kumar2020active}, which fail to capture the temporal dynamics inherent in image streams. As a result, they frequently miss critical learning opportunities or expend resources on redundant data. More advanced theoretical frameworks, such as Shapley values \cite{ghorbani2019data} or influence functions \cite{koh2017understanding}, require extensive retraining and are computationally prohibitive for edge deployment.

\begin{figure}[t]
    \centering
    \includegraphics[width=0.95\columnwidth]{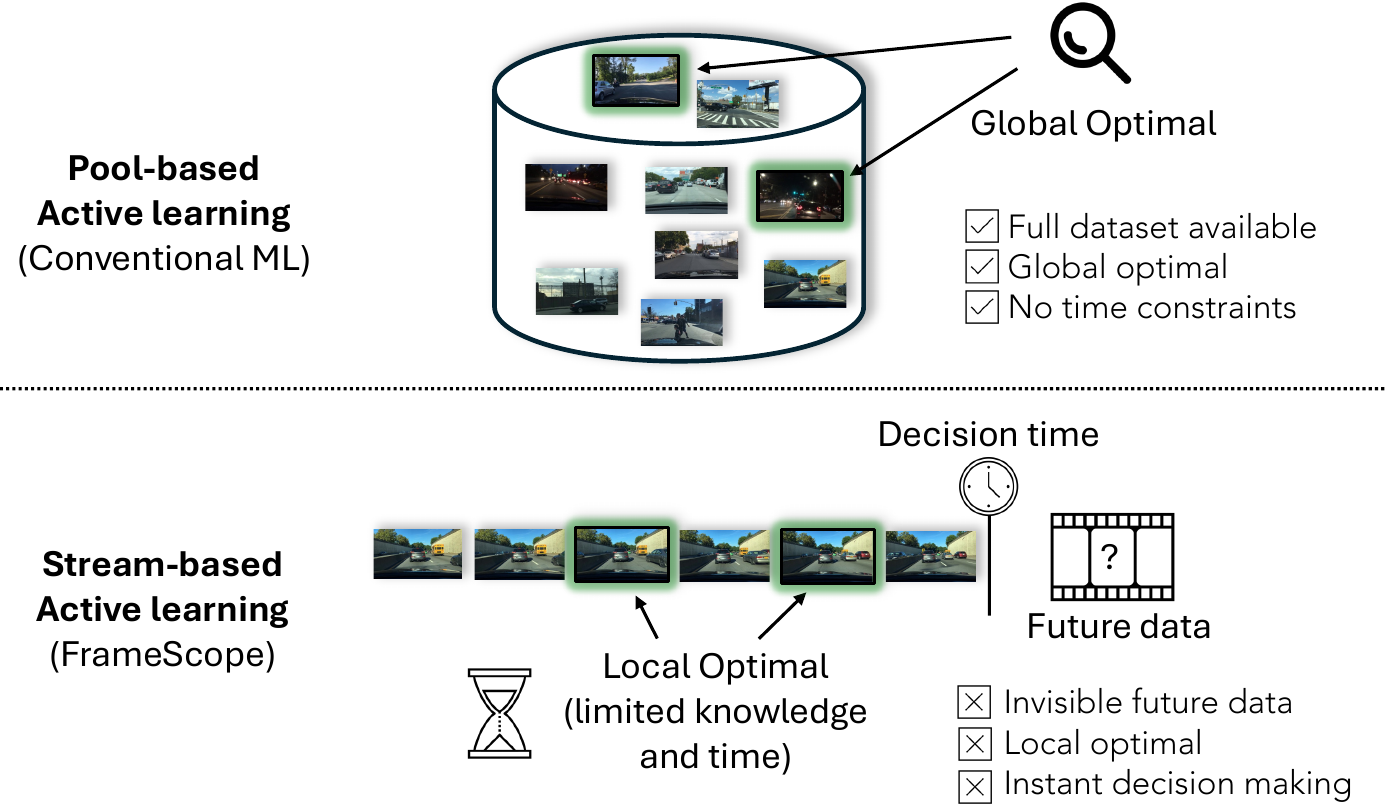}
    \vspace{-0.2cm}
    \caption{Traditional pool-based active learning (top), such as DAVINZ~\cite{wu2022davinz} versus stream-based active learning (bottom), as in \ourmethod{}.}
    \vspace{-0.4cm}
    \label{fig:moti}
\end{figure}

Recent advances in data valuation theory offer promising solutions. The DAVINZ framework~\cite{wu2022davinz} demonstrated that neural tangent kernel (NTK) theory~\cite{jacot2018neural} can estimate data importance for model performance without requiring full model retraining, providing computational advantages over gradient-based~\cite{koh2017understanding} and retraining-based~\cite{ghorbani2019data} methods. However, DAVINZ is a \emph{pool-based} active learning approach designed for static datasets. As illustrated in Figure~\ref{fig:moti}, pool-based methods assume full access to the dataset \emph{a priori} to enable global optimization of sample selection without time constraints. These methods assess data value based on each instance's contribution to overall model performance, treating all samples as independent. Accordingly, they have primarily been applied to static image classification datasets such as CIFAR-10~\cite{krizhevsky2009learning} and MNIST~\cite{deng2012mnist}.

However, continuous learning in autonomous driving operates under an \emph{stream-based} active learning paradigm, where the assumptions of the pool-based methods break down. In stream-based settings, visual data arrives sequentially, and \emph{future} frames are unavailable at the time of data valuation and selection. This streaming nature fundamentally alters the problem formulation: the value of a frame must be assessed in real time, without knowledge of what comes next. This constraint necessitates local optimization under limited information and demands full exploitation of available spatio-temporal cues.

To address these challenges, we introduce \ourmethod, a framework that leverages data valuation theory in the temporal domain to enable effective continuous learning in autonomous vehicle vision systems. At its core is the \emph{Temporal Neural Tangent Kernel (TNTK)}, which offers both theoretical and algorithmic advancements over the classical NTK formulation. TNTK captures temporal dependencies across image frames that are temporally adjacent by incorporating two key innovations: (1) a temporal position weighting scheme which emphasizes contextually rich center frames; and (2) scene dynamics weighting, which adaptively prioritizes high-dynamics frames that are more likely to capture critical driving maneuvers. We realize this via a temporal data valuation core integrated with a lightweight adaptation loop on the vehicle.

We validate our approach through comprehensive experiments across multiple domain shift scenarios—geographic, weather, and time—using open-source datasets. The results demonstrate consistent performance improvements over baseline methods across different sampling rates and continuous learning phases. In addition to achieving superior final performance, our method offers greater sampling efficiency and effectively mitigates catastrophic forgetting. While existing data valuation methods in autonomous driving rely on heuristic strategies~\cite{zhu2024adaptav}, metadata-based selection~\cite{trinh2024data}, or computationally intensive influence functions~\cite{li2024data}, our work pioneers the translation of NTK theory to the valuation of streaming video data. To the best of our knowledge, this is the first work to introduce NTK-based data valuation within the context of continuous learning for autonomous vehicle vision systems.

We also present a complete \emph{on-vehicle model adaptation system architecture} that integrates data valuation, query selection, and continuous learning in a closed-loop manner. It consists of three key components: (i) frame selection via temporal data valuation, (ii) cloud-based oracle model for high-quality label generation, and (iii) on-vehicle GPU-based model retraining. By transmitting only selected high-value frames to the cloud, the system ensures bandwidth and computational efficiency suitable for edge deployment, while preserving continuous adaptation capabilities.

The implications of this work extend beyond autonomous driving to any domain that requires continuous learning from streaming visual data, including surveillance systems, robotics, and online video analysis. By providing a theoretically grounded, yet computationally efficient approach to temporal data valuation, our method addresses a fundamental challenge in deploying machine learning systems in dynamic, real-world environments.

\section{Related Work}

\subsection{Data Valuation}

\noindent\textbf{Traditional data valuation methods.}
Data valuation quantifies individual data points' contribution to model performance, crucial for fair resource allocation. Early approaches rooted in cooperative game theory utilize the Shapley value \cite{shapley1953value} for fair utility distribution among contributing players. Ghorbani and Zou \cite{ghorbani2019data} formalized Data Shapley for supervised learning, where data points serve as players and model performance as a utility function, satisfying efficiency, symmetry, and additivity properties. However, exact Shapley value calculation requires exponential complexity across all possible data subsets. Jia et al.~\cite{jia2019towards} developed approximation algorithms including Truncated Monte Carlo and Group Testing, making Shapley-based data valuation practical for real-world applications.

\vspace{0.25\baselineskip}
\noindent\textbf{Deep learning-based data valuation.}
Traditional methods requiring extensive retraining became computationally prohibitive for large-scale deep learning. Wu et al.~\cite{wu2022davinz} introduced DAVINZ, a training-free approach leveraging Neural Tangent Kernel (NTK) theory and Maximum Mean Discrepancy (MMD) to estimate data value without model convergence. By exploiting neural network properties at initialization and deriving domain-aware generalization bounds, DAVINZ reduces computation from hours to minutes while maintaining correlation with ground truth valuations and theoretical properties of Shapley-based methods.

\subsection{Active Learning and Sample Selection}

\noindent\textbf{Traditional active learning approaches.}
Active learning reduces annotation costs by strategically selecting informative unlabeled examples \cite{settles2009active, cacciarelli2025active}. Main strategies include uncertainty sampling (selecting the highest prediction uncertainty examples) \cite{lewis1994heterogeneous, yang2018benchmark}, query-by-committee (employing multiple models to identify prediction disagreement) \cite{seung1992query}, and expected model change (selecting examples causing the greatest parameter changes) \cite{settles2008analysis, roy2001toward}.

\vspace{0.25\baselineskip}
\noindent\textbf{Deep learning-based active learning.}
Modern neural networks introduce computational efficiency and representation learning challenges \cite{ash2019deep}. Advances focus on Bayesian approaches like Monte Carlo Dropout \cite{gal2017deep} and BALD \cite{houlsby2011bayesian} for uncertainty estimation, and diversity-based methods like Core-Set \cite{sener2017active} and BADGE \cite{ash2019deep} combine uncertainty with representativeness to avoid redundant selection.

\vspace{0.25\baselineskip}
\noindent\textbf{Streaming and continuous active learning.}
Stream-based active learning must contend with challenges such as concept drift, limited storage, and real-time constraints. Continuous active learning extends this paradigm to handle evolving data distributions and tasks~\cite{rios2024cual}, requiring a balance between exploring new concepts and exploiting existing knowledge. AdaptAV~\cite{zhu2024adaptav} proposed a cloud-based continuous learning framework for autonomous vehicles that leverages cloud-hosted oracle models for ground truth generation and model retraining. While demonstrating the feasibility of cloud-assisted learning, AdaptAV relies on heuristic frame sampling strategies, such as periodic, density-based, and event-based methods, without employing principled data valuation. This approach may miss critical learning opportunities or expend resources on suboptimal samples. 

Hawk~\cite{george2023low} integrates semi-supervised, active, and transfer learning into a live learning pipeline that performs bandwidth-aware, selective transmission to remote labelers under tight constrained backhaul. It is primarily optimized for efficient rare-positive acquisition through bandwidth-adaptive ranking and serves as a strong, well-engineered baseline for edge-based continuous improvement. Similar to Hawk, but unlike cloud-centric approaches that transmit all frames to the cloud for sampling and retraining, \ourmethod{} selects high-value frames directly on-vehicle and requests labels only for those. The key distinction lies in the \emph{selection criterion}: \ourmethod{} performs temporally-aware, distribution-shift-robust data valuation, whereas Hawk emphasizes rare-positive recovery via bandwidth-adaptive ranking. Accordingly, \ourmethod{} aims for continual calibration under evolving conditions, complementing Hawk's focus on efficient rare-event collection under extreme scarcity.
\section{\ourmethod{} System Architecture}

Deciding where to execute the continuous learning loop introduces system‑level constraints. A cloud‑only retraining architecture would repeatedly push large model weights to the fleet—straining downlink budgets (e.g., SSD‑MobileNet $\approx$ 30 MB; Faster R‑CNN $\approx$ 160 MB) and require the cloud service to track per‑vehicle model state (versions, checkpoints, routing), while centralizing updates for many vehicles creates scaling and queueing pressure. These considerations motivate performing data valuation and retraining on the vehicle. 
However, processing 100\% of streaming video data for on-vehicle model retraining imposes prohibitive computational and bandwidth demands, making selective sampling essential for viable active learning in autonomous vehicles. These challenges are compounded by mobile network conditions, where fluctuating bandwidth and transmission delays make continuous data upload impractical. At the same time, on-vehicle edge computing resources are far more limited than those in the cloud; processing or temporarily storing large volumes of video data can quickly exhaust available capacity. These constraints necessitate intelligent, lightweight sampling decisions, as opposed to the complex optimization procedures feasible only in pool-based active learning. Furthermore, the dynamic nature of driving environments leads to temporal decay in data value, as the usefulness of training samples diminishes over time with changing conditions, routes, and scenarios.

\begin{figure}[t]
    \centering
    \includegraphics[width=1\columnwidth]{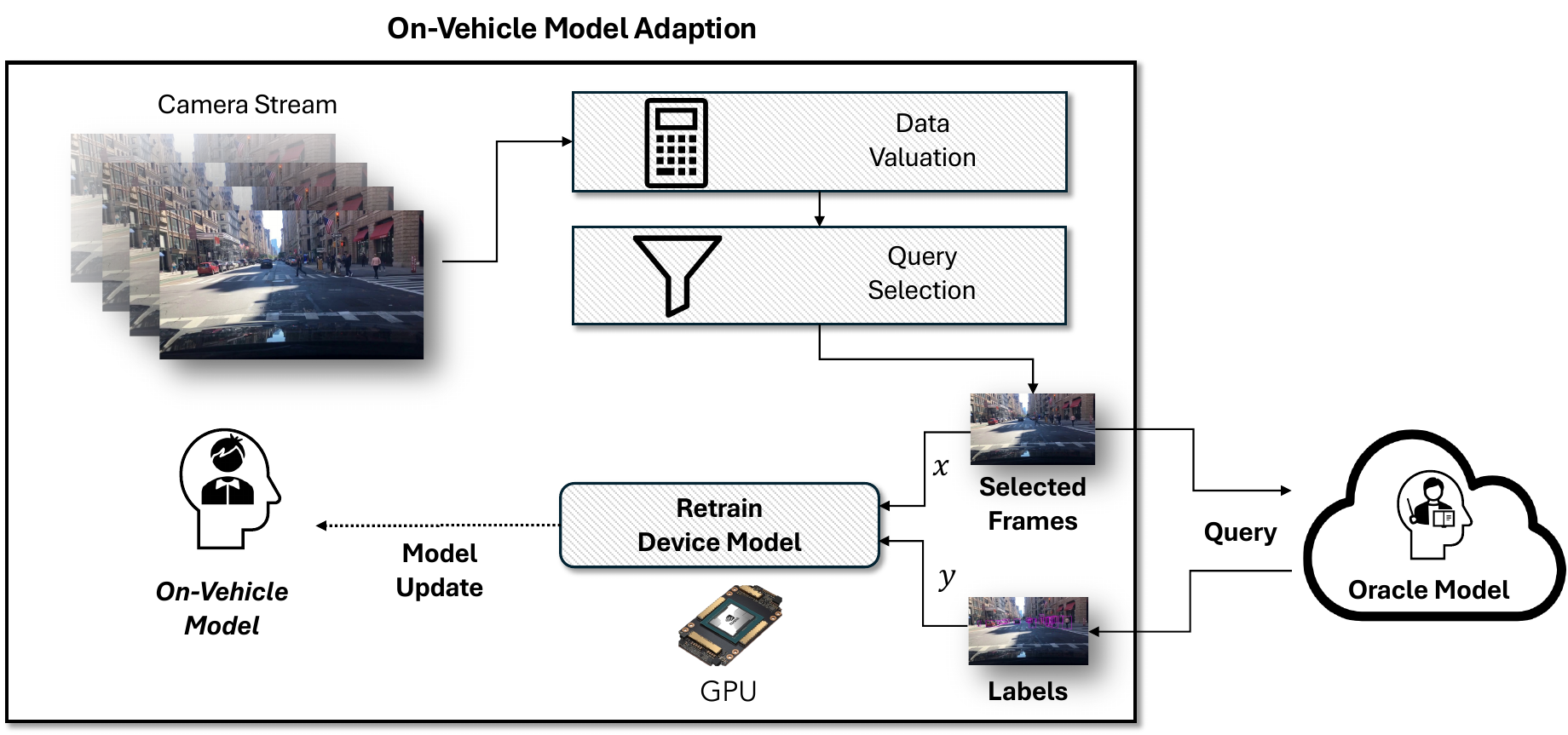}
    \caption{\ourmethod{} system architecture.}
    \label{fig:sys_diag}
\end{figure}

\ourmethod{} enables on-vehicle model adaptation that continuously improves autonomous driving perception through selective data valuation and query-based learning. It introduces principled data valuation with theoretically grounded frame selection and executes the end-to-end loop on the vehicle while using the cloud only as a labeling oracle for queried frames. This design enables edge deployment under constrained bandwidth and computes budgets without centralizing sampling or retraining.

The overall architecture, illustrated in Figure~\ref{fig:sys_diag}, comprises three main components operating in a closed-loop manner:

\vspace{0.2\baselineskip}
\noindent\textbf{Camera Stream}: The system continuously receives real-time video streams from vehicle-mounted cameras, which capture dynamic driving scenarios. 

\vspace{0.2\baselineskip}
\noindent\textbf{On-Vehicle Model Adaptation}: It orchestrates the frame valuation and selection and model updating process through three interconnected modules:
\begin{itemize}
    \item \textit{Data Valuation Module}: Employs our proposed \ourmethod\ score to evaluate the learning value of incoming camera frames, considering both visual complexity and temporal dependencies.
    \item \textit{Query Selection Module}: Acts as a filter that selects the most valuable frames for annotation based on the computed data valuation scores and available labeling budget constraints.
    \item \textit{Model Retraining Module}: Utilizes on-vehicle GPU resources to update the perception model using newly labeled high-value frames.
\end{itemize}
\noindent\textbf{Oracle Model}: The system leverages cloud-based oracle models with superior accuracy and computational resources to provide high-quality ground-truth labels for the queried frames, which enables knowledge transfer to the resource-constrained on-vehicle model.

\vspace{0.5\baselineskip}
Model retraining runs opportunistically (e.g., during idle windows) and is warm‑started from the current model. The loop targets condition‑level drift, such as illumination and weather, and a small replay buffer helps preserve prior competencies. This update policy reduces model weight transfers and keeps the cloud stateless.

\section{Stream Data Valuation using Temporal Neural Tangent Kernel}\label{sec:methodology}

In this section, we present our temporal data valuation method designed for stream-based active learning in autonomous driving scenarios.

\subsection{Problem Statement and Contribution}

Originally designed for static datasets, DAVINZ faces the following challenges when applied directly to video streams:
\begin{itemize}
    \item \emph{Temporal blindness}: It fails to capture inter-frame dependencies critical for understanding motion and context in video.
    \item \emph{Granularity mismatch}: Dataset-level valuation is not directly applicable to frame-level valuation required in streaming scenarios.
    \item \emph{Dynamic scene complexity}: Static assumptions cannot handle motion and scene dynamics.
\end{itemize}
To address the above challenges, we propose the following innovations:
\begin{itemize}
    \item \emph{Temporal Neural Tangent Kernel (TNTK)}: extends NTK theory to temporal domains.
    \item \emph{Frame-level MMD (FL-MMD) computation}: refines domain variance analysis to single-frame granularity.
    \item \emph{Multi-scale timing modeling}: adapts to the dynamics of different time scales.
\end{itemize}

\subsection{Problem Definition}

Given a continuously arriving video stream $\mathcal{S} = \{\mathbf{x}_1, \mathbf{x}_2, ..., \mathbf{x}_i, ...\}$, where each frame $\mathbf{x}_i$ represents a multi-dimensional visual data, the goal is to learn a selection function (i.e., whether to include a frame in the retraining set) $\phi: \mathcal{F} \rightarrow [0,1]$ that maximizes the model performance improvement under a labeled budget constraint $B$: 
\begin{equation}
    \phi^* = \arg\max_{\phi} \mathbb{E}\left[\sum_{\mathbf{x} \in \mathcal{S}} \phi(\mathbf{x}) \cdot \Delta\mathcal{L}(\mathbf{x})\right] \quad \text{s.t.} \quad \sum_{\mathbf{x} \in \mathcal{S}} \phi(\mathbf{x}) \leq B,
\end{equation}
where $\Delta\mathcal{L}(\mathbf{x})$ denotes the marginal contribution of frame $\mathbf{x}$ to the model performance, and $B$ is the labeling budget. The model performance can be quantified by, for example, object detection accuracy, semantic segmentation IoU (Intersection over Union), classification error rates, etc. The labeling budget represents the maximum number of frames that can be sent to the cloud oracle for labeling, given the computational and bandwidth constraints.

\begin{figure}[t]
    \centering
    \includegraphics[width=1\columnwidth]{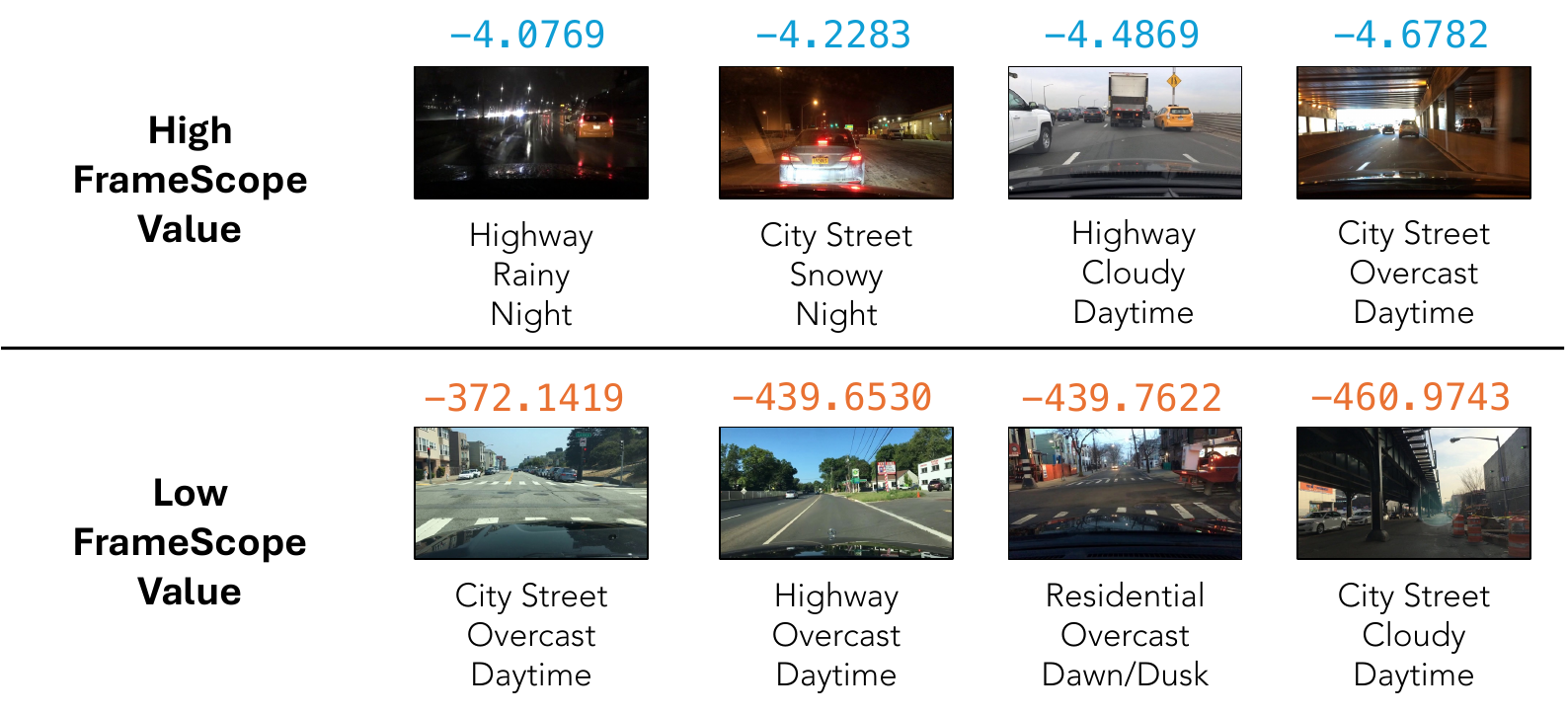}
    \caption{Example of high \ourmethod\ value (top) and low \ourmethod\ value (bottom) frames. 
    } 
    \label{fig:high_low_value}
\end{figure}

Our goal is to select the most valuable frames for retraining under budget constraints, \emph{without} observing the entire dataset. This is a fundamental constraint in real-world autonomous driving scenarios, where future visual data is unavailable. To address this, we introduce a data valuation metric called the \emph{\ourmethod{} score}. It is designed to capture both temporal dependencies among frames and scene dynamics. Figure~\ref{fig:high_low_value} illustrates examples of high- and low-value frames, where high-value frames are expected to contribute more significantly to model performance improvement. The score estimates training benefit in temporal context. Domain labels do not directly determine the score; novelty relative to prior observations and representativeness within the local window are primary factors. In the remainder of this section, we describe how to compute the \ourmethod{} score for each frame in a continuous video stream and how it is used to select high-value frames for continuous learning.

\subsection{Preliminary: DAVINZ Framework}

The DAVINZ framework evaluates the \emph{learning value} of data, specifically quantifying how much a given dataset $\mathcal{S}$ can contribute to model performance improvement through two core components: 
\begin{equation}\label{eq:DAVINZ}
    \nu(\mathcal{S}) = -\kappa\sqrt{\frac{\hat{y}^T\Theta_0^{-1}\hat{y}}{m_{\mathcal{S}}}} - d_{\mathcal{H}}(\mathcal{T},\mathcal{S}),
\end{equation}
where $\hat{y}$ represents the model predictions on dataset $\mathcal{S}$, $\Theta_0$ is the Neural Tangent Kernel (NTK) matrix at initialization, $m_{\mathcal{S}}$ denotes the size of dataset $\mathcal{S}$, $\kappa$ is a scaling parameter, and $d_{\mathcal{H}}(\mathcal{T},\mathcal{S})$ measures the Maximum Mean Discrepancy (MMD) between candidate domain distribution $\mathcal{T}$ and source dataset distribution $\mathcal{S}$ in reproducing kernel Hilbert space $\mathcal{H}$. The first component captures the generalization bound (i.e., how good the data contributes to training) based on the neural tangent kernel theory, while the second component quantifies the domain distribution alignment. $\nu(S)$ is always negative, and the closer it is to zero, the higher the value of the given dataset $S$ is.

To understand how DAVINZ quantifies the learning value through the first component, we briefly review the Neural Tangent Kernel theory that underlies its generalization bound. Neural Tangent Kernel theory provides a theoretical framework for understanding the training dynamics of deep neural networks by characterizing how the network function changes during gradient descent training. The theory reveals that the training dynamics of an infinitely wide neural network can be inscribed by fixing the kernel function: 
\begin{equation}\label{eq:ntk}
    \Theta(\mathbf{x},\mathbf{x}') = \nabla_\theta f(\mathbf{x},\theta_0)^T \nabla_\theta f(\mathbf{x}',\theta_0),
\end{equation}
where $f(\mathbf{x},\theta)$ represents the neural network function with parameters $\theta$, and $\nabla_\theta f(\mathbf{x},\theta_0)$ is the gradient of the network output with respect to the parameters at initialization. That is, it characterizes the sensitivity of the network's output to changes in the model parameters for a given input $\mathbf{x}$. Accordingly, Eq.~\eqref{eq:ntk} quantifies the similarity between the effects of two different inputs, $\mathbf{x}$ and $\mathbf{x}'$, on the network during gradient descent training.

DAVINZ constructs its domain-aware generalization bound based on the NTK theory. However, in continuous video streams, temporal dependencies between frames carry critical semantic information that classical NTK theory does not capture. For instance, in autonomous driving scenarios, the value of a sharp turn is determined not only by the current frame but also by the broader motion trajectory formed by preceding and subsequent frames. Traditional NTK treats each frame as an independent sample and thus overlooks the temporal structure essential for understanding such dynamic events.

\begin{figure*}[htb]
    \centering
    \includegraphics[width=0.75\textwidth]{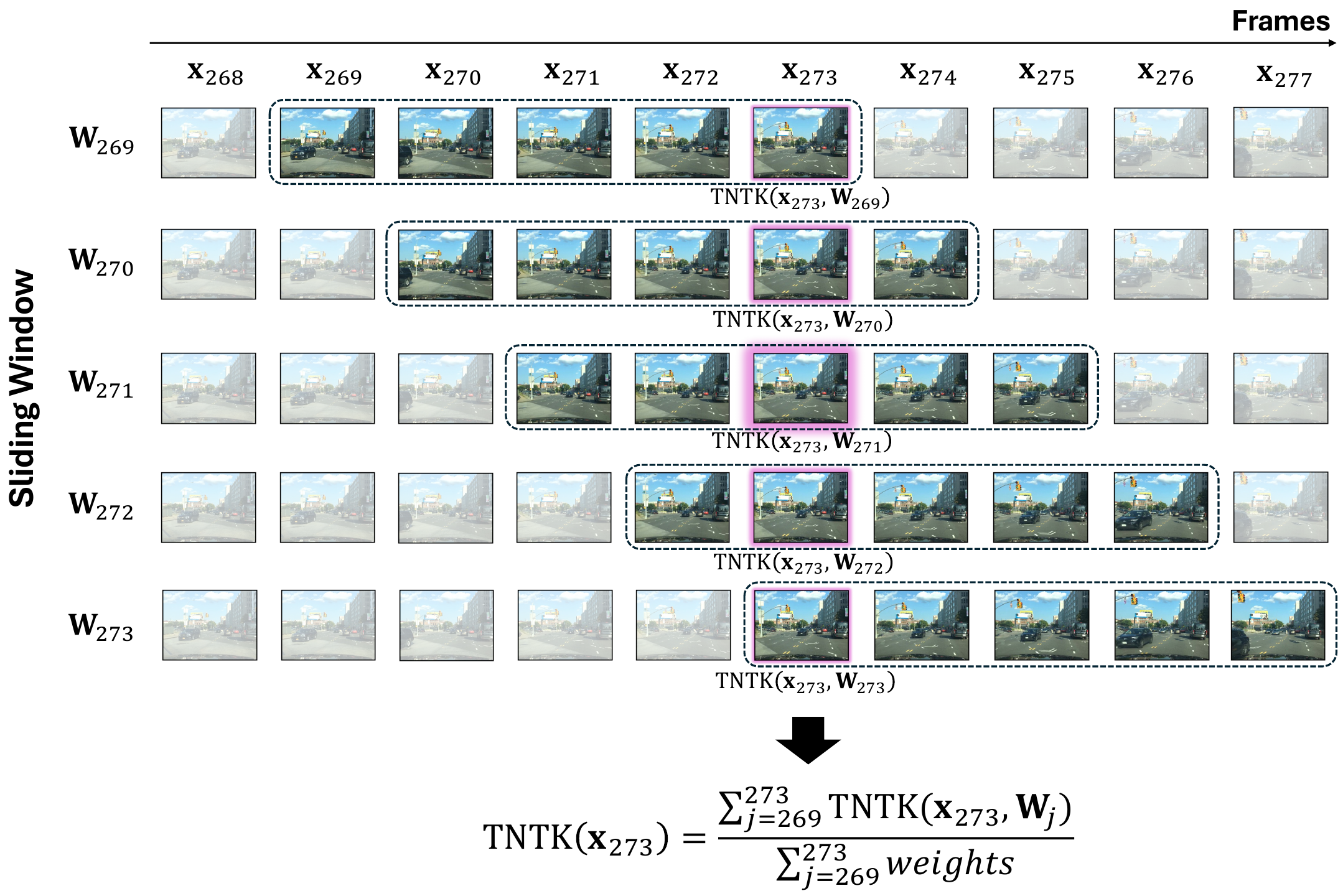}
    \caption{\ourmethod{}'s frame-wise TNTK calculation using window-wise NTK scores. A frame's TNTK value is assessed based on its temporal position within multiple overlapping sliding windows.}
    \label{fig:high-level}
\end{figure*}

\vspace{0.5\baselineskip}

\subsection{Temporal Extension of DAVINZ Components}

To address the limitations of directly applying DAVINZ to video streams, we propose temporal extensions to both components of the DAVINZ scoring function in Eq.~\eqref{eq:DAVINZ}:

\begin{enumerate}
    \item \emph{Temporal Neural Tangent Kernel (TNTK)}: We extend the first component $-\kappa\sqrt{\frac{\hat{y}^T\Theta_0^{-1}\hat{y}}{m_{\mathcal{S}}}}$ (in-domain complexity) to incorporate temporal dependencies through sliding window analysis and frame-level weighting.
    
    \item \emph{Frame-Level Maximum Mean Discrepancy (FL-MMD)}: We refine the second component, $d_{\mathcal{H}}(\mathcal{T},\mathcal{S})$, by transitioning from dataset-level to frame-level domain discrepancy computation.
\end{enumerate}
With the above temporal extensions, we can define the \ourmethod\ score for each frame in a continuous video stream:
\begin{definition}[\ourmethod\ Score] \label{def:temporal_davinz}
For a frame $\mathbf{x}_i$ in a continuous video stream, its \ourmethod\ score is defined as follows:
\begin{equation}\label{eq:temporal_davinz}
    \nu(\mathbf{x}_i) = -\text{TNTK}(\mathbf{x}_i) - d_{\mathcal{H}}(\mathcal{T},\mathbf{x}_i),
\end{equation}
where $\text{TNTK}(\mathbf{x}_i)$ is the Temporal Neural Tangent Kernel score that captures the temporal-aware in-domain complexity, and $d_{\mathcal{H}}(\mathcal{T},\mathbf{x}_i)$ represents the frame-level domain discrepancy. Algorithm~\ref{alg:temporal_davinz} summarizes the computation steps for the \ourmethod\ score. The detailed steps for computing the TNTK and FL-MMD are provided in Sections~\ref{sec:tntk} and~\ref{sec:frame_mmd}, respectively.
\end{definition}

\begin{algorithm}[t]
\caption{\ourmethod\ Score for Frame $\mathbf{x}_i$, $\nu(\mathbf{x}_i)$}
\label{alg:temporal_davinz}
\begin{algorithmic}[1]
\Require Video stream $\mathcal{S}$, frame $\mathbf{x}_i$, target domain $\mathcal{T}$, window size $w$, decay parameter $\lambda$, threshold $\theta$

\State \textbf{Step 1:} Compute TNTK score
\State $tntk\_score = \text{TNTK}(\mathbf{x}_i)$ \Comment{Section~\ref{sec:tntk}}
\vspace{0.5\baselineskip}
\State \textbf{Step 2:} Compute frame-level MMD score  \Comment{Section~\ref{sec:frame_mmd}}
\State Extract features: $\mathbf{feat}_i = f(\mathbf{x}_i)$ using current on-vehicle model
\State Extract target features: $\{\mathbf{feat}_{t_k}\}_{k=1}^{|\mathcal{T}|} = \{f(\mathbf{t}_k)\}_{k=1}^{|\mathcal{T}|}$
\State $fl\_mmd\_score = \text{FL-MMD}(\mathbf{feat}_i, \{\mathbf{feat}_{t_k}\}_{k=1}^{|\mathcal{T}|})$ 

\vspace{0.5\baselineskip}
\State \textbf{Step 3:} Combine both components
\State $\nu(\mathbf{x}_i) = -tntk\_score - fl\_mmd\_score$ \Comment{Eq.~\eqref{eq:temporal_davinz}}

\vspace{0.5\baselineskip}
\State \Return $\nu(\mathbf{x}_i)$ 
\end{algorithmic}
\end{algorithm}

\begin{figure*}[t]
    \centering
    \includegraphics[width=0.75\textwidth]{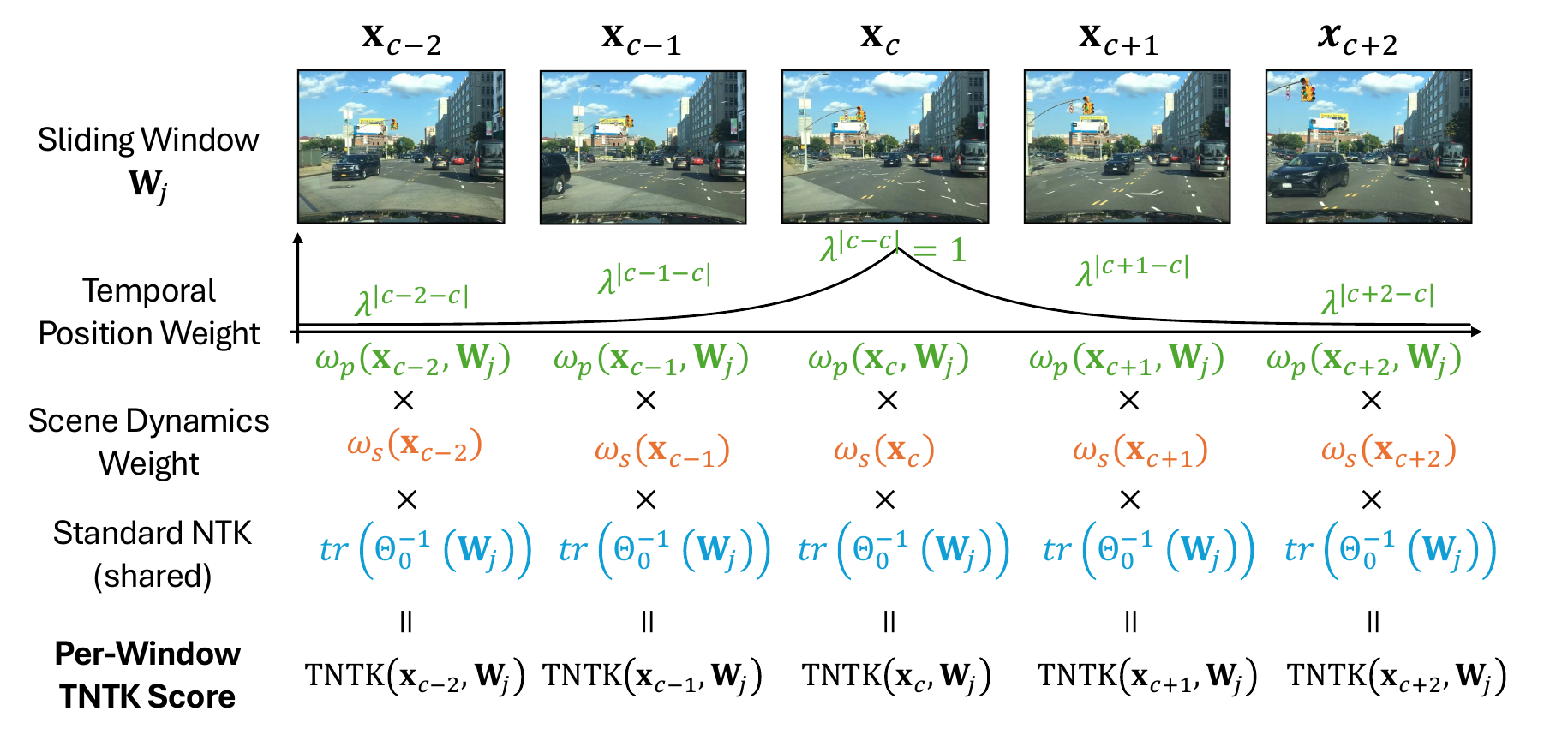}
    \caption{The per-window TNTK calculation for frames in a sliding window.}
    \label{fig:TNTK_diagram}
\end{figure*}

\subsection{Temporal Neural Tangent Kernel}\label{sec:tntk}

\noindent\textbf{High-Level Idea:}
To address the set-based valuation limitations of the standard NTK theory, we introduce a novel Temporal Neural Tangent Kernel (TNTK) method that enables frame-wise valuation in continuous video streams. The key idea is to compute the NTK values over overlapping sliding windows of frames. This allows us to capture temporal dependencies while still enabling frame-level valuation.

Figure~\ref{fig:high-level} illustrates the high-level idea of our TNTK method. In this figure, we show a continuous video stream with frames $\mathbf{x}_{268}, \mathbf{x}_{269}, \ldots, \mathbf{x}_{277}$, and multiple sliding windows $\{\mathbf{W}_j\}$ of size $w=5$ that overlap with each other. As we can see, each frame $\mathbf{x}_i$ appears in multiple overlapping windows with different temporal positions: for instance, frame $\mathbf{x}_{273}$ appears in windows $\mathbf{W}_{269}$, $\ldots$, $\mathbf{W}_{273}$. Within each window, all frames share a single NTK value (as the standard NTK theory does support set-level valuation only), which represents the \emph{collective learning value} of that temporal window as a whole. However, our key insight is that different frames within a window contribute differently to this collective value based on their temporal positions. For example, the center frame $\mathbf{x}_{273}$ in window $\mathbf{W}_{271}$ has richer bi-directional temporal context than boundary frames like $\mathbf{x}_{271}$ or $\mathbf{x}_{275}$. But, in the other windows, the frame $\mathbf{x}_{273}$ may not be the most informative frame. Thus, our valuation method operates under the key assumption that center frames in a sliding window can best represent the window's temporal context.

To achieve frame-level valuation using window-level NTK values, we employ a weighted aggregation strategy. Each frame receives two weights: \emph{Temporal Position Weight}, which is based on its relative position within each sliding window, with center frames receiving higher weights, and \emph{Scene Dynamics Weight}, which reflects the amount of visual dynamics between consecutive frames. The final score for a frame is then computed as a weighted average of its contributions across all sliding windows that contain it, where the weights are determined by both the temporal position and scene dynamics. See Figure~\ref{fig:high-level} again for the illustration. For frame $\mathbf{x}_{273}$, its per-window TNTK value is computed as $\text{TNTK}(\mathbf{x}_{273}, \mathbf{W}_j)$ for each sliding window $\mathbf{W}_j$ that contains it. The final TNTK score for $\mathbf{x}_{273}$ is then computed as the weighted average of these per-window scores. This design enables each frame to be evaluated from multiple temporal perspectives while addressing the computational constraints of NTK theory.

\vspace{0.5\baselineskip}
\noindent\textbf{Detailed Steps:}
We first define the \emph{per-window} TNTK score for frame $\mathbf{x}_i$ within a single sliding window $\mathbf{W}_j$ (which is illustrated in Figure~\ref{fig:TNTK_diagram}):
\begin{equation}\label{eq:tntk}
    \text{TNTK}(\mathbf{x}_i,\mathbf{W}_j) = tr(\Theta_0^{-1}(\mathbf{W}_j)) \cdot \omega_p(\mathbf{x}_i, \mathbf{W}_j) \cdot \omega_s(\mathbf{x}_i),
\end{equation}
where $tr(\Theta_0^{-1}(\mathbf{W}_j))$ is the trace of the inverse NTK matrix computed over all frames within window $\mathbf{W}_j$, capturing the collective learning complexity of the window based on the neural tangent kernel theory. It essentially quantifies the amount of gradient information that the temporal window can contribute to model training. The Temporal Position Weight $\omega_p(\mathbf{x}_i, \mathbf{W}_j)$ reflects the relative importance of frame $\mathbf{x}_i$ within window $\mathbf{W}_j$, assigning higher weights to frames near the center of the window due to their richer bidirectional temporal context. In parallel, the Scene Dynamics Weight $\omega_s(\mathbf{x}_i)$ captures the importance of frame $\mathbf{x}_i$ based on the magnitude of visual changes between consecutive frames, prioritizing dynamic scenes over static ones. Together, these two weighting mechanisms ensure that the most temporally and contextually informative frames receive appropriate emphasis in the final TNTK computation. We formally define both weights shortly below.

Once we compute the per-window TNTK score for frame $\mathbf{x}_i$ across all sliding windows $\mathbf{W}_j$ that contain it, we can aggregate these scores to obtain the final TNTK score for frame $\mathbf{x}_i$. Specifically, we compute the weighted average of the per-window TNTK scores:
\begin{equation}
    \text{TNTK}(\mathbf{x}_i) = \frac{\sum_{j} \text{TNTK}(\mathbf{x}_i, \mathbf{W}_j)}{\sum_{j} \omega_p(\mathbf{x}_i, \mathbf{W}_j) \cdot \omega_s(\mathbf{x}_i)}.
\end{equation}
Algorithm~\ref{alg:tntk} summarizes the TNTK scoring operation.

\vspace{0.5\baselineskip}

\begin{algorithm}[tb]
\caption{TNTK Score for Frame $\mathbf{x}_i$, $\text{TNTK}(\mathbf{x}_i)$}
\label{alg:tntk}
\begin{algorithmic}[1]
\Require Frame $\mathbf{x}_i$, sliding windows containing $\mathbf{x}_i$: $\{\mathbf{W}_j \mid \mathbf{x}_i \in \mathbf{W}_j\}$, window size $w$, decay parameter $\lambda$, scene dynamics threshold $\theta$

\State Initialize $total\_score = 0$, $total\_weight = 0$
\For{each sliding window $\mathbf{W}_j$ containing $\mathbf{x}_i$}
    \State $\boldsymbol{\Theta}_0(\mathbf{W}_j) = [\nabla_{\boldsymbol{\theta}} f(\mathbf{x}_k,\boldsymbol{\theta}_0)^T \nabla_{\boldsymbol{\theta}} f(\mathbf{x}_l,\boldsymbol{\theta}_0)]_{k,l}$
    \State $kernel\_trace = tr(\boldsymbol{\Theta}_0^{-1}(\mathbf{W}_j))$ \Comment{Trace approximation of $\hat{y}^T\Theta_0^{-1}\hat{y}/m_S$}
    
    \State $\mathbf{x}_c$: center frame of window $\mathbf{W}_j$ 
    \State $\omega_p = \lambda^{|i-c|}$ \Comment{Temporal position weight}
    \State $\omega_s = \text{Algorithm~\ref{alg:msw}}(\mathbf{x}_i)$ \Comment{Scene dynamics weight}
    
    \State $\text{TNTK}(\mathbf{x}_i,\mathbf{W}_j) = kernel\_trace \times \omega_p \times \omega_s$
    \State $total\_score += \text{TNTK}(\mathbf{x}_i,\mathbf{W}_j)$
    \State $total\_weight += \omega_p \times \omega_s$
\EndFor

\State $\text{TNTK}(\mathbf{x}_i) = total\_score / total\_weight$ \Comment{Normalization}
\State \Return $\text{TNTK}(\mathbf{x}_i)$
\end{algorithmic}
\end{algorithm}

\vspace{0.5\baselineskip}

\noindent\textbf{Temporal position weight:} 
It quantifies the influence of frame position within a temporal sliding window using an exponential decay function:
\begin{equation}\label{eq:decay_param}
\omega_p(\mathbf{x}_i, \mathbf{W}_j) = \lambda^{|i - c|},    
\end{equation}
where $c$ is the index of the window $\mathbf{W}_j$'s center frame $\mathbf{x}_c$, and $\lambda$ is the decay parameter. 
The design of temporal position weight employs an exponential decay mechanism that naturally emphasizes central frames while progressively reducing the influence of temporally distant frames. 
This function exhibits the following key characteristics:
\begin{itemize}
    \item \emph{Center maximization}: When $i = c$, $\omega_p(\mathbf{x}_c, \mathbf{W}_j) = \lambda^0 = 1$, ensuring maximum weight for the center frame.
    \item \emph{Distance monotonicity}: Weights decrease monotonically with distance from the center, with faster decay for smaller decay parameter $\lambda$, preserving temporal locality.
    \item \emph{Position sensitivity}: Each frame receives different weights across different sliding windows depending on its relative position within each window.
    \item \emph{Computational efficiency}: The simple exponential form enables efficient computation on edge devices.
\end{itemize}

\vspace{0.5\baselineskip}
\noindent\textbf{Scene dynamics weight:} 
In our TNTK model, frames with higher scene dynamics are weighted more heavily based on the following intuitions:
\begin{itemize}
    \item \emph{Rare event capture}: High scene dynamics often indicate unusual or challenging scenarios such as emergency braking, sharp turns, or obstacle avoidance that are underrepresented in training data but critical for safety, making them more valuable for learning.
    \item \emph{Temporal complexity}: Dynamic scenes contain richer temporal dependencies and spatial transformations and thus provide more diverse training targets compared to static scenes.
    \item \emph{Redundancy reduction}: Static or low-motion scenes often exhibit temporal redundancy where consecutive frames contain nearly identical content, hence they lead to diminishing returns in model training. Including such redundant frames may also slow down training convergence and waste computational resources without providing new learning signals, whereas high-dynamics frames offer more diverse and informative training samples.
\end{itemize}

\begin{definition}[Scene Dynamics] \label{def:scene_dynamics}
Given consecutive frames $\mathbf{x}_i$ and $\mathbf{x}_{i+1}$ with their latent feature representations 
$f(\mathbf{x}_i)$ and $f(\mathbf{x}_{i+1})$ extracted through the current on-vehicle model,\footnote{In our implementation based on SSD\_MobileNet, we extract features from the \texttt{base\_net} component of the model, which comprises the MobileNet feature extraction layers. The features are obtained by applying global average pooling over the spatial dimensions to produce a 1024-dimensional feature vector for each input frame.} the scene dynamics is defined as:
\begin{equation}
    M(\mathbf{x}_i) = 1 - \mathrm{cos\_sim}\big(f(\mathbf{x}_i), f(\mathbf{x}_{i+1})\big),\label{eq:scene_dynamics}
\end{equation}
where $\mathrm{cos\_sim}(\mathbf{u}, \mathbf{v})=\frac{\mathbf{u}\cdot \mathbf{v}}{||\mathbf{u}|| ||\mathbf{v}||}$ denotes the cosine similarity function of two vectors $\mathbf{u}$ and $\mathbf{v}$. 
\end{definition}

\begin{algorithm}[t]
\caption{Scene Dynamics Weight $\omega_s(\mathbf{x}_i)$}
\label{alg:msw}
\begin{algorithmic}[1]
\Require Frame index $i$, scene dynamics threshold $\theta$, feature extractor $f(\cdot)$
    \State Extract features: $\mathbf{feat}_i = f(\mathbf{x}_i)$, $\mathbf{feat}_{i+1} = f(\mathbf{x}_{i+1})$
    \State Compute cosine similarity: $\mathrm{cos\_sim} = \frac{\mathbf{feat}_i \cdot \mathbf{feat}_{i+1}}{||\mathbf{feat}_i|| \cdot ||\mathbf{feat}_{i+1}||}$
    \State $M_i = 1 - \mathrm{cos\_sim}$ \Comment{Scene dynamics}
    \If{$M_i > \theta$}
        \State $\omega_s = 1 + \frac{M_i}{\theta}$ \Comment{Higher weight due to dynamic scene}
    \Else
        \State $\omega_s = 1$ \Comment{Baseline weight}
    \EndIf
\State \Return $\omega_s$
\end{algorithmic}
\end{algorithm}

\begin{figure*}[t]
    \centering
    \includegraphics[width=1\textwidth]{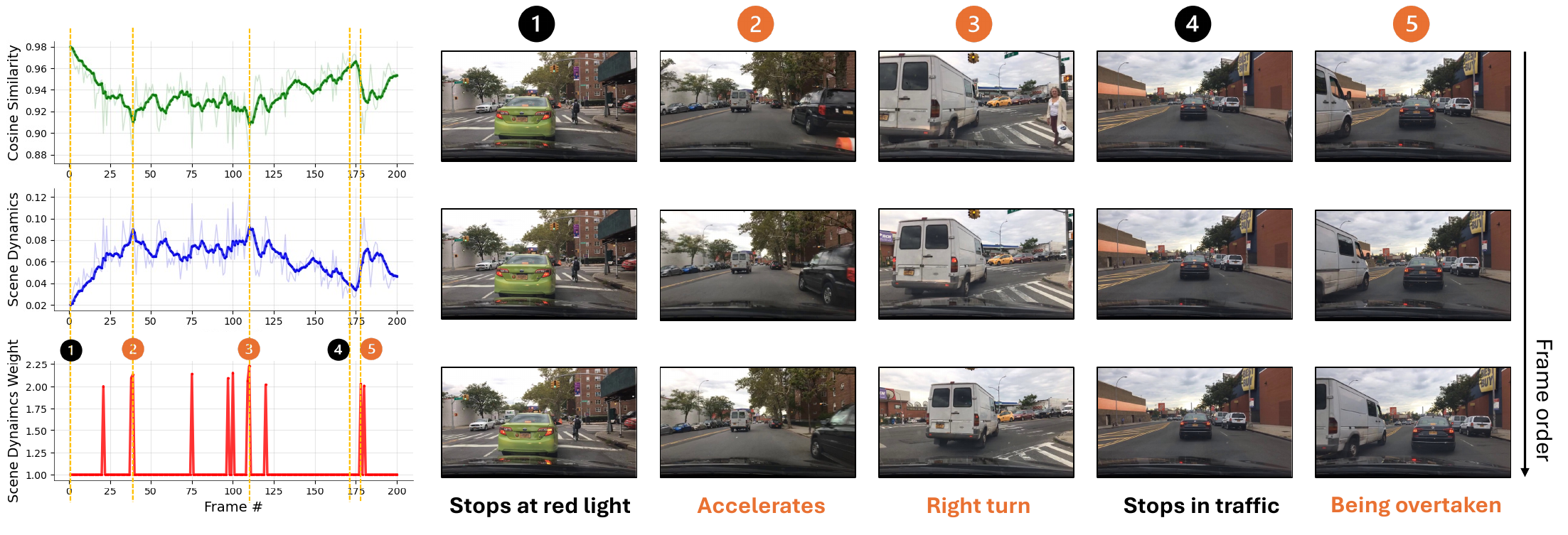}
    \vspace{-0.5cm}
    \caption{An example illustrating how the Scene Dynamics Weight varies across different driving scenarios within a single driving sequence.}
    \label{fig:scene_dynamics_exapmle}
\end{figure*}

We use cosine similarity because it can capture directional changes in the latent feature space, which reflects semantic or structural differences rather than mere pixel-level variations. By comparing the angle between normalized feature vectors, it highlights meaningful content changes while being robust to variations in feature magnitudes caused by factors like lighting or scale. This makes it a reliable metric for detecting scene dynamics, as large angular differences typically correspond to significant visual changes. As a result, Eq.~\eqref{eq:scene_dynamics} serves as a practical and interpretable proxy for scene dynamics.

\vspace{0.5\baselineskip}
\begin{definition}[Scene Dynamics Weight] \label{def:scene_dynamics_weight}
Based on the scene dynamics, the Scene Dynamics Weight is defined as an adaptive weight function:
\begin{equation}\label{eq:scene_dynamics_weight}
    \omega_s(\mathbf{x}_i) = \begin{cases}
    1 + \frac{M(\mathbf{x}_i)}{\theta} & \text{if } M(\mathbf{x}_i) > \theta, \\
    1 & \text{otherwise,}
    \end{cases}
\end{equation}
where $\theta$ (e.g., 0.1) is the scene dynamics threshold parameter used to filter low-intensity noise movements.
\end{definition}

Eq.~\eqref{eq:scene_dynamics} quantifies inter-frame dissimilarity, while Eq.~\eqref{eq:scene_dynamics_weight} provides an adaptive mechanism to capture dynamic variations in information value. Traditional temporal modeling often assumes that all frames possess identical information density. However, in dynamic visual scenarios such as autonomous driving, different frames exhibit varying degrees of variation in information value. The Scene Dynamics Weight is specifically designed to address this challenge based on our core hypothesis that inter-frame scene dynamics is positively correlated with the potential learning value for model training. This assumption is grounded in the maximum entropy principle from information theory; high-dynamics regions typically correspond to higher information entropy, thus possessing greater learning value.

The adaptive weight function in Eq.~\eqref{eq:scene_dynamics_weight} is designed to balance multiple considerations: it preserves a baseline by assigning a minimum weight of 1 to all frames, which prevents excessive downweighting of low-dynamics frames that may still contain valuable information, such as challenging object appearances or subtle weather conditions. Simultaneously, the function amplifies high-dynamics frames—those exceeding the threshold—by assigning additional weight, reflecting their potentially greater learning value. Its linear scaling property ensures that weights grow proportionally with scene dynamics, effectively capturing the continuous nature of scene adaptation. Finally, the threshold-based design provides upper-bound control and thus prevents excessive weight inflation that could destabilize the learning process.

\vspace{0.5\baselineskip}
\noindent\textbf{Example: } 
Figure~\ref{fig:scene_dynamics_exapmle} illustrates how the Scene Dynamics Weight varies across different driving scenarios within a 200-frame sequence. The graphs on the left display three synchronized curves: cosine similarity (top), scene dynamics (middle), and the resulting Scene Dynamics Weight (bottom). The Scene Dynamics Weight curve reflects the threshold-based filtering mechanism: weights remain at the baseline value of 1.0 when scene dynamics fall below the threshold $\theta$ (which we set to 0.1), effectively filtering out low-dynamics scenes while amplifying significant dynamic events. By adjusting $\theta$, users can control the sensitivity of the system. The image samples on the right highlight five driving scenarios that demonstrate the expected behavior of our metric: 
\begin{enumerate}
    \item[(1)] When the ego vehicle stops at a red light, consecutive frames exhibit high cosine similarity, leading to low Scene Dynamics Weight, which indicates minimal learning value.
    \item[(2)] As the ego vehicle accelerates after passing an intersection, cosine similarity decreases and Scene Dynamics Weight increases accordingly.
    \item[(3)] Significant visual changes during a right turn result in higher scene dynamics and elevated Scene Dynamics Weight, capturing the increased learning potential of such maneuvers.
    \item[(4)] The metrics return to values similar to those in scenario (1) as the ego vehicle stops for traffic, again indicating a static scene.
    \item[(5)] Most notably, when a white minivan quickly passes the stationary ego vehicle from the side, cosine similarity drops sharply and the Scene Dynamics Weight peaks. This rare and dynamic event is correctly identified as highly valuable for training.
\end{enumerate}
This example demonstrates that the Scene Dynamics Weight mechanism assigns appropriately higher weights to dynamic events than to static scenes, aligning with their relative learning value.

\begin{table*}[t]
\caption{Number of video records for different domains.}
\label{table:domain_shift}
\centering
\footnotesize
\begin{tabular}{crrrrrrrrrrrr}
\toprule
\multirow{2}{*}{\textbf{Domain}}     &
\multicolumn{4}{|c|}{\textbf{Geographic}}              & \multicolumn{5}{|c|}{\textbf{Weather}}            & \multicolumn{3}{|c}{\textbf{Time}}    \\

& \multicolumn{1}{|r}{highway} & residential & city street & \multicolumn{1}{r|}{other} & clear & cloudy & rainy & snowy & \multicolumn{1}{r|}{foggy} & daytime & dawn/dusk & night \\
\midrule
\textbf{Record\#} & 52      & 19          & 106         & 3     & 95    & 34     & 16    & 16    & 19    & 96      & 19        & 65   \\
\bottomrule
\end{tabular}
\end{table*}

\subsection{Frame-level Domain Discrepancy}\label{sec:frame_mmd}

While TNTK addresses the temporal extension of the in-domain complexity term, we also need to adapt the second component of DAVINZ, i.e., the domain discrepancy term $d_{\mathcal{H}}(\mathcal{T},\mathcal{S})$, for frame-level evaluation. In the original DAVINZ framework, MMD is computed between entire datasets. However, for stream-based active learning, we require frame-level domain discrepancy scores.

For a frame $\mathbf{x}_i$ in a continuous video stream, its frame-level domain discrepancy, FL-MMD, is defined as follows:
\begin{equation}\label{eq:frame_mmd}
    d_{\mathcal{H}}(\mathcal{T},\mathbf{x}_i) = \text{MMD}(f(\mathbf{x}_i), \{f(\mathbf{t}_k)\}_{k=1}^{|\mathcal{T}|}),
\end{equation}
where $f(\mathbf{x}_i)$ denotes the feature representation of frame $\mathbf{x}_i$ extracted using the current on-vehicle model, and $\{f(\mathbf{t}_k)\}_{k=1}^{|\mathcal{T}|}$ represent the feature embeddings of target domain samples. This frame-level MMD quantifies how dissimilar an individual frame is from the target domain distribution, enabling fine-grained assessment of domain alignment.

\section{Evaluation}

\subsection{Experiment Setup}
We use a workstation equipped with an Intel i9-13900F CPU and an NVIDIA GeForce RTX 4090 GPU. Although the GPU has 24 GB of video memory by default, we manually limit its maximum VRAM usage to 4 GB to simulate the constrained computing resources typical of edge platforms.

\subsection{Datasets}
\begin{itemize}
    \item Waymo Open Dataset~\cite{Sun_2020_CVPR_Waymo}: 
    It serves as the \textit{source domain} in our domain shifting experiments; that is, the on-vehicle model is initially trained on this dataset. For the initial weight training, we choose the first 180 video records (each has about 200 frames) from the validation set in Waymo Open Dataset v1.4.2. It is captured at 10 Hz with 1920x1280 resolution. 

    \item BDD100K~\cite{bdd100k}: It serves as the \textit{target domain} in our domain shifting experiments; that is, the on-vehicle model has never seen this dataset before. But, it is used for continuous learning after the initial model (trained on the Waymo Dataset) is deployed on the vehicle. We use the MOT-2020 object detection set that consists of 200 dashcam video sequences, each of which has about 200 frames, resulting in a total of 39,973 frames. It is captured at 5 Hz in 1280x720 resolution. 
\end{itemize}

\subsection{Models}
\begin{itemize}
    \item SSD\_MobileNet~\cite{liu2016ssd, howard2017mobilenets}: It is the \emph{on-vehicle model} that is subject to continuous learning. It is an object detection model that combines Single Shot Multibox Detector with a MobileNet architecture, making it fast and efficient for real-time applications on edge devices. It contains 3,206,976 parameters, and the weight file size is 27,636 KB. It is important to note that we choose this model not only because it enables real-time object detection in vehicles, but also because of the correlation between the model parameters and the size of the training set in continuous learning. 
    \item Faster R-CNN~\cite{ren2015faster}: It serves as an \emph{Oracle model} providing labeling information upon each query sent from the vehicle. Faster R-CNN is a high-performance object detection model. It is utilized with a region proposal network to proposes candidate regions, and a second stage classifies them and refines the bounding boxes. The model contains 41,092,136 parameters, more than 12 times the number of parameters in SSD\_MobileNet.
\end{itemize}

\begin{figure*}[t]
    \centering
    \vspace{-0cm}
    \includegraphics[width=0.95\textwidth]{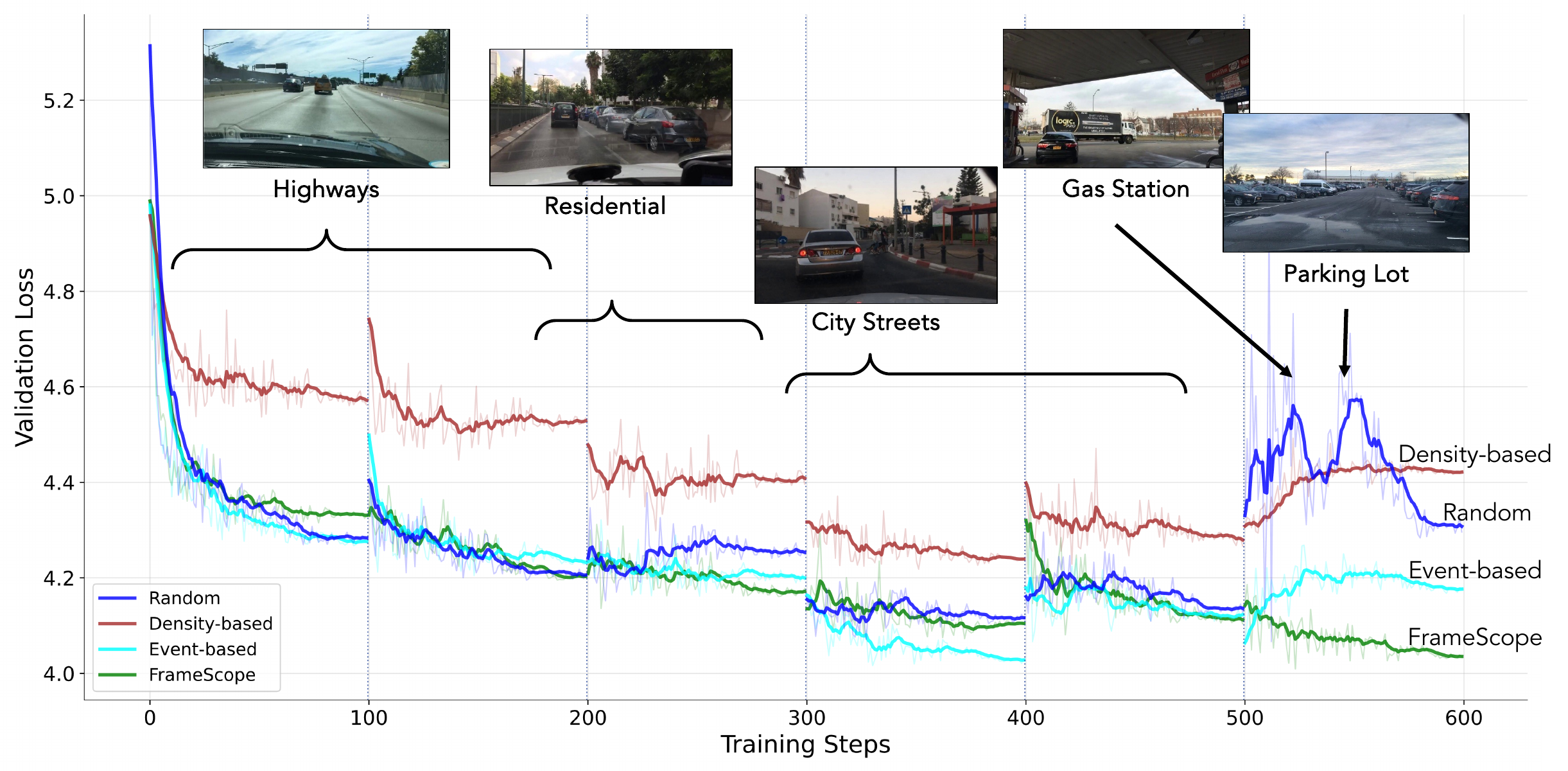}
    \vspace{-0cm}
    \caption{Continuous learning performance across geographic domain transitions. The system encounters sequential domain shifts from highways to residential areas, city streets, gas stations, and parking lots. Our \ourmethod{} approach (green) consistently outperforms baseline methods, demonstrating superior adaptation to new environments while maintaining stability throughout the domain transitions.}
    \label{fig:geographic_loss}
    \vspace{-0cm}
\end{figure*}

\subsection{Results}

In this section, we answer the following questions:
\begin{itemize}
    \item How does our \ourmethod{} sampling perform compared to baseline approaches under different domain-shifting scenarios?
    \item What is the impact of sampling rate on model performance and resource efficiency?
    \item What are the computational and bandwidth advantages of our edge-based sampling approach compared to the cloud-centric approach?
\end{itemize}

\vspace{0.5\baselineskip}
\noindent\textbf{Domain shifting.}
The metadata that accompanies the BDD100K dataset, such as geographic location, weather condition, and temporal context (time of day), provides valuable information about the driving environment, which we leverage to create domain-shifting scenarios. The driving environment is split into three main domains: geographic, weather, and time. The distribution of each condition for three domains is shown in Table~\ref{table:domain_shift}.

Out of 200 video records in the BDD100K dataset, we use 180 records for the continuous learning. These are the video sequences that an autonomous vehicle would observe during actual road driving, from which our sampling methods will select frames for model retraining. We arrange these 180 records with different environmental conditions in order, roughly from easy to hard. For example, as shown in Figure~\ref{fig:geographic_loss}, the geographic domain follows an ordering that begins with highways, followed by residential areas and then city streets, reflecting a progression in environmental complexity. Finally, the last records contain even more difficult scenarios, such as non-road environments like gas stations and parking lots. The order of domain shifting is consistent with the order reflected in Table~\ref{table:domain_shift}. After arranging the records by difficulty, we split the 180 records into six groups of 30 records each, representing six consecutive retraining phases. This setup simulates the natural progression of an autonomous vehicle encountering increasingly complex driving scenarios over time. 
Note that a single environment condition spans multiple phases. For example, the highway condition appears in the first two phases, while the city street condition appears in the last four phases. This arrangement allows us to evaluate how well the model adapts to new scenarios while retaining knowledge from previous conditions.

The remaining 20 records are used as a validation set, which is held out to evaluate the model's performance after each retraining phase. The validation set is sampled in equal proportion to the total dataset distribution, ensuring that the distribution across domains matches the distribution of the entire dataset.
When continuous learning is performed effectively with appropriate sample selection, the model should adapt smoothly to new domain conditions while retaining knowledge from previous scenarios, which will show lower validation loss. Conversely, poor sample selection may lead to catastrophic forgetting of earlier domains or inadequate adaptation to new, challenging scenarios.

\vspace{0.5\baselineskip}
\noindent\textbf{Sampling methods and implementation details.}
To evaluate the effectiveness of our proposed \ourmethod{} sampling approach, we compare it against the following baseline methods adapted from existing literature and established practices in active learning for autonomous driving scenarios:
\begin{itemize}
    \item \emph{Random sampling} serves as the fundamental baseline. It uniformly selects frames without any intelligent criteria.
    \item \emph{Event-based sampling} \cite{zhu2024adaptav} selects frames when significant changes occur in the scene (e.g., appearance or disappearance of objects).
    \item \emph{Density-based sampling}, also adopted from \cite{zhu2024adaptav}, prioritizes frames with higher object density as determined by the oracle model predictions (thus requires cloud-side sampling).
\end{itemize}
In all experiments, we apply a consistent sampling rate across methods. The \emph{sampling rate} denotes the percentage of frames selected from each group of image records for oracle annotation and model retraining. For instance, at a 10\% sampling rate, approximately 600 frames with the highest \ourmethod{} scores are selected from each retraining phase with 6,000 frames, while the remaining 90\% are discarded without processing. This constraint reflects practical resource limitations in autonomous vehicle deployments, where querying and annotation costs and communication bandwidth are major constraints.

\vspace{0.5\baselineskip}
\emph{\ourmethod{} scoring implementation:} 
We set the sliding window size to $w=5$ frames, with a stride of $\lfloor w/2 \rfloor = 2$ to balance contextual richness and computational efficiency. The temporal decay factor is configured as $\lambda = 0.8$. For computing Scene Dynamics Weight, we use a threshold of $\theta = 0.1$. The \ourmethod{} score is computed according to the formulation described in Section~\ref{sec:methodology}. We compute scores over overlapping windows to retain temporal context but issue label queries at the frame level to avoid clip‑level labeling and uplink overhead.

To mitigate forgetting in continual learning, we implement an \emph{experience replay} mechanism alongside our \ourmethod{} sampling strategy. The replay buffer stores high-value frames from previous phases based on their \ourmethod{} scores. During the retraining phase $p$, the training set combines new samples from the current phase with replay samples from earlier phases $i < p$ in a fixed 70:30 ratio. The 30\% replay portion is allocated across previous phases using a temporal decay weight $w_i = \gamma^{p - i}, \quad i = 1, \ldots, p-1$, where $\gamma \in (0,1)$ controls the decay rate. These weights are normalized to yield sampling ratios:
\[
r_i = \frac{w_i}{\sum_{j=1}^{p-1} w_j}, \quad n_i = N_{\text{replay}} \times r_i,
\]
with each $n_i$ sample drawn randomly from phase $i$'s high-value frames. For example, at phase $p=4$ with $\gamma=0.7$, the weights are $w_1=0.343, w_2=0.49, w_3=0.7$, normalizing to ratios $r_1 \approx 0.22, r_2 \approx 0.32, r_3 \approx 0.46$. With $N_{\text{replay}} = 180$, this yields about 39, 58, and 83 replay samples from Phases 1, 2, and 3, respectively. This stratified replay strategy ensures temporal diversity while emphasizing more recent experiences, which helps preserve prior knowledge and improve adaptation to new domain conditions.

\vspace{0.5\baselineskip}
\noindent\textbf{Computational complexity and memory requirement.} The filtering loop is training‑free: each frame undergoes one forward feature extraction, followed by closed‑form temporal aggregation over a sliding window. The amortized per‑frame time complexity is $O(d+w)$: the $O(d)$ term comes from feature‑level operations proportional to the feature dimension $d$ (e.g., dot products, cosine similarity, or simple transforms) that are performed once per frame; the $O(w)$ term comes from aggregating over a window of length $w$ using cached features (constant‑time per position, summed over $w$ positions). Because features are computed once and reused while the window slides, we do not re‑compute features $w$ times per frame. The working memory scales as $O(wd)$ for maintaining a rolling buffer of $w$ feature vectors of dimension $d$.

\vspace{0.5\baselineskip}
\noindent\textbf{Continuous learning performance.}
Figure~\ref{fig:geographic_loss} shows the validation loss progression throughout the six-phase geographic domain shifting experiment with a 10\% sampling rate. Each training phase comprises 100 training steps as indicated by the vertical dashed lines at steps 100, 200, 300, 400, and 500. While experiments were performed for all domain shift cases, we present only the geographic domain results here due to space constraints. All three domain shifts exhibited similar trends; we chose to highlight the geographic domain because the changes in scene context are more visually distinct compared to the weather and time domains, and it allows us to showcase clearly different scenarios, such as gas stations and parking lots.

The y-axis represents the validation loss computed on a separate validation set held out during training, where lower values indicate better model performance and generalization. The x-axis shows the cumulative retraining steps across all six phases, during which progressively more challenging driving scenarios are introduced. The curves reveal several important insights into the effectiveness of different sampling strategies in continuous learning. Initially, all methods start with comparable validation loss, but their trajectories diverge significantly as training progresses through increasingly complex driving scenarios.

Our \ourmethod{} demonstrates the most stable and consistent performance improvement, achieving the lowest final validation loss among all compared approaches. It exhibits smooth convergence without significant performance degradation during domain transitions. This indicates effective mitigation of catastrophic forgetting.

Event-based and density-based sampling methods show moderate performance but exhibit more fluctuations during domain transitions. While these methods achieve reasonable final results, they lack the stability demonstrated by our \ourmethod{}, particularly during the middle training phases where performance plateaus occur. Event-based sampling reacts only to immediate visual changes and tends to miss gradual domain shifts. For instance, when transitioning from highway to city environments, it may skip important frames that do not trigger object appearance events but contain valuable adaptation information. Density-based sampling relies solely on object counts, which fails to capture the visual complexity necessary for effective domain adaptation. It cannot distinguish between frames containing familiar objects in new contexts and those that provide truly informative learning cues. In contrast, our \ourmethod{} method evaluates each frame's learning potential through gradient analysis while incorporating temporal context, resulting in more effective frame selection across domain boundaries.

Interestingly, random sampling exhibits a distinctive pattern characterized by a sudden performance degradation in the final retraining phase, as shown by the sharp increase in validation loss between steps 500 and 550. This observation supports our hypothesis that the final training group contains complex and uncommon non-road scenarios, such as gas stations and parking lots. These challenging scenarios likely cause instability in random sampling due to the absence of strategic selection criteria when encountering novel situations that differ significantly from previously seen data.

\begin{table}[tb]
\caption{AP@50 (car) under a 10\% sampling rate after six continual phases.
Values are mean AP@50~$\pm$~std. The last row shows 95\% confidence intervals for the paired AP@50 difference between \ourmethod{} and the best non-\ourmethod{} baseline in each scenario.}
\centering
\footnotesize
\begin{tabular}{llll}
\toprule
\textbf{Sampling Method} & \textbf{Geographic} & \textbf{Time} & \textbf{Weather} \\
\toprule
Random        & 0.476$\pm$0.0028 & 0.472$\pm$0.0028 & 0.506$\pm$0.0028 \\
Density-based & 0.484$\pm$0.0031 & 0.434$\pm$0.0034 & 0.491$\pm$0.0029 \\
Event-based   & 0.504$\pm$0.0029 & 0.469$\pm$0.0041 & 0.509$\pm$0.0029 \\
\textbf{\ourmethod{}}  & \textbf{0.529$\pm$0.0026} & \textbf{0.494$\pm$0.0027} & \textbf{0.519$\pm$0.0028} \\
\midrule
95\% CI for $\Delta$ & [$0.022$, $0.028$] & [$0.019$, $0.026$] &  [$0.006$, $0.012$] \\
\bottomrule
\end{tabular}
\label{table:final_results}
\end{table}

\vspace{0.25\baselineskip}

Table~\ref{table:final_results} presents the final detection accuracy obtained after six continual training phases across three domain-shifting scenarios on BDD100K. Specifically, it shows AP@50 (Average Precision at IoU = 0.50) for the \textit{car} class, expressed as mean~$\pm$~standard deviation. AP values are rounded to three decimals and standard deviations to four decimals to reflect empirical uncertainty. 
The bottom row reports the 95\% confidence intervals for the \emph{paired} AP differences, $\Delta=\text{AP}(\ourmethod{})-\text{AP}(\text{best baseline})$, computed via paired bootstrap on the test set (bootstrap replicates $B{=}1,000$, test set size $N{=}4,000$). An interval crossing zero would imply the difference is indistinguishable from evaluation noise. Moreover, the farther the entire interval is from zero, the stronger the evidence that the difference is statistically significant.

The results in Table~\ref{table:final_results} show that \ourmethod{} is consistently superior across all three domain-shifting scenarios. This outperformance is statistically significant, as the 95\% confidence intervals for the paired AP differences are strictly positive. Specifically, it achieves the highest AP@50 scores in the geographic (0.529), time (0.494), and weather (0.519) domains, representing significant improvements over the best-performing baseline methods in each case.
Event-based sampling performs well in the geographic and weather scenarios, achieving strong results that outperform density-based and random sampling methods. However, its performance drops significantly in the time domain (0.479), likely because gradual lighting changes fail to trigger sufficient event-based selections. Density-based sampling shows inconsistent performance across domains, with particularly poor results in the time scenario (0.445). These findings suggest that object density alone is not a reliable indicator of frame informativeness under varying environmental conditions.

These results validate our hypothesis that incorporating temporal dynamics and scene dynamics through the \ourmethod{} framework leads to more robust sample selection. The consistent performance gains across diverse domain-shifting scenarios highlight the generalizability of our approach and its practical value for real-world autonomous driving applications, where vehicles encounter varied and continuously evolving environmental conditions.

\vspace{0.5\baselineskip}
\noindent\textbf{Bandwidth requirement.}
An important practical consideration is the bandwidth requirement of different continuous learning strategies. Both event-based and density-based (“\textit{Filter-on-Cloud}”) baselines rely on the entire image set to make sampling decisions: event-based sampling tracks object entries and exits across the sequence, while density-based sampling counts objects in every frame. When scoring is performed in the cloud, all frames must be transmitted, which inflates network usage and cost. In contrast, \ourmethod{} makes sampling decisions locally using model-internal signals within short windows. Hence, only selected high-value frames are transmitted for oracle annotation. 

\begin{table}[t]
\centering
\caption{Per-phase communication cost (6,000 frames, 10\% selection).}
\label{tab:bandwidth-merged}
\footnotesize
\begin{tabular}{l lrr} 
\toprule
\textbf{Category} & \textbf{Model/Method} & \textbf{Uplink} & \textbf{Downlink} \\
\toprule
\multirow{2}{*}{\textbf{Train-on-Cloud}} & SSD\_MobileNet & 15.5\,GB & 27.6\,MB (weights) \\
 & Faster R-CNN & 15.5\,GB & 159.7\,MB (weights) \\
\midrule
\multirow{2}{*}{\textbf{Filter-on-Cloud}} & Density-based & 15.5\,GB & 1.9\,MB (labels) \\
 & Event-based & 15.5\,GB & 1.3\,MB (labels) \\
 \midrule
\textbf{Our Method} & \textbf{\ourmethod{}} & \textbf{1.5\,GB} & \textbf{0.97\,MB (labels)} \\
\bottomrule
\end{tabular}
\end{table}

Under a common per-phase budget (6,000 encountered frames with a 10\% sampling rate, i.e., 600 labeled samples), Table~\ref{tab:bandwidth-merged} quantifies the resulting uplink and downlink communication volumes. The cloud-training baselines ("\textit{Train-on-Cloud}") upload 100\% of frames and download updated model weights once per phase (initial weights are a one-time cost and excluded here). The "\textit{Filter-on-Cloud}" methods also upload all frames for oracle's scoring and filtering but download only the labels of the post-filtered samples. Note that model weight updates are typically 30-160x larger than label payloads, depending on the model. Our \ourmethod{} performs filtering on-vehicle, uploading only the selected frames (\textbf{90.3\%} fewer bytes), and downloading only the corresponding labels. Consequently, it substantially reduces both uplink and downlink communication volumes, making it better suit deployments with constrained connectivity.

\begin{figure}[tb]
    \centering
    \includegraphics[width=1\linewidth]{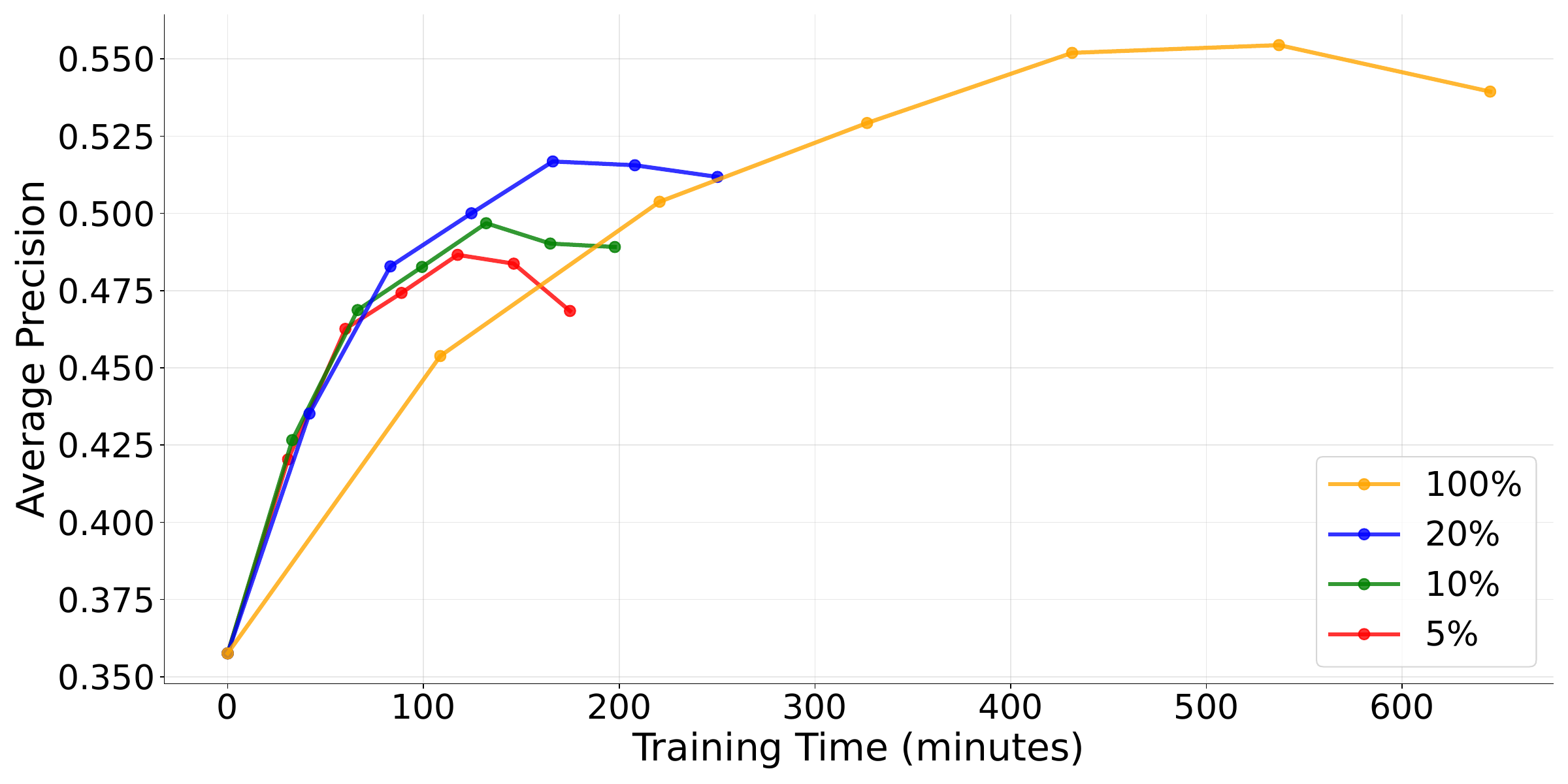}
    \caption{Training curves comparing different sampling rates (5\%, 10\%, 20\%, 100\%) using \ourmethod{} under time domain shifting. 
    }
    \label{fig:sampling_rates_time}
\end{figure}

\vspace{0.5\baselineskip}
\noindent\textbf{Sampling rate analysis and efficiency considerations.}
To investigate the trade-off between annotation cost (measured by the number of frames selected for oracle labeling and model retraining) and model performance, we analyze our \ourmethod{} method under varying sampling rates in the time domain shifting scenario. Specifically, we compare sampling rates of 5\%, 10\%, 20\%, and 100\% (the full dataset baseline). Figure~\ref{fig:sampling_rates_time} shows the training progression curves, illustrating the average precision on the BDD100K car class over six continuous training phases.

The results reveal several key insights into the trade-offs between sampling efficiency and resource utilization in continuous learning scenarios. Most notably, all sub-sampling rates (5\%, 10\%, and 20\%) exhibit similar convergence behavior during the initial training phases. This suggests that our \ourmethod{} method effectively identifies the most informative frames, even under tight annotation budgets. The stability of early-stage performance supports the theoretical foundation of our approach—that the high-value frames selected by \ourmethod{} contain sufficient information to enable effective initial model adaptation.

As expected, higher sampling rates lead to better final performance, with the 100\% baseline achieving the highest average precision of approximately 0.552. However, the 20\% sampling rate yields competitive results with an average precision of 0.522—only a 4.7\% drop—while requiring 80\% fewer samples and substantially lower bandwidth consumption. Even more notably, the 10\% sampling rate maintains reasonable performance at 0.494, demonstrating that our method can achieve roughly 90\% of the full-dataset performance using just 10\% of the data.

From a practical deployment perspective, these results highlight the necessity and benefits of principled sampling strategies in resource-constrained environments. Bandwidth requirements scale linearly with the sampling rate—moving from 5\% to 20\% sampling increases network transmission costs by a factor of four, while using the full 100\% dataset leads to a 20-fold increase.
Our 10\% sampling approach offers a proper balance by achieving about 90\% of the full-dataset performance while reducing bandwidth consumption by roughly an order of magnitude, thus serving as a practical default for deployment.

Furthermore, the superior early-stage performance of our proposed sampling method addresses a fundamental challenge in autonomous vehicle deployment: the need for rapid model adaptation to new environmental conditions. In scenarios such as entering unfamiliar geographic regions or encountering sudden weather changes, the ability to quickly adapt using minimal data while maintaining competitive performance offers a critical advantage over methods that rely on extensive data collection and processing.

\section{Conclusion}

In this paper, we presented \ourmethod, the first system to extend neural tangent kernel theory to the temporal domain for data valuation in autonomous vehicle systems. Our approach addresses key limitations of existing methods that rely on heuristic sampling and fail to capture the temporal dynamics inherent in video streams. We introduced the Temporal Neural Tangent Kernel (TNTK) framework, which incorporates both temporal dependencies among frames and scene dynamics to enable principled frame selection. Comprehensive experimental results demonstrated the practical utility of our approach in real-world settings, particularly under bandwidth and computational constraints. Future work will explore adaptive hyperparameter tuning, multi-modal integration, and cross-domain applications. Overall, the \ourmethod{} framework marks a step toward bridging theoretical data valuation and practical streaming applications and establishes a new paradigm for intelligent data selection in continuous learning scenarios.

\bibliographystyle{ACM-Reference-Format}
\bibliography{refs}

\end{document}